\documentclass[letterpaper, 10 pt, conference]{ieeeconf}  

\IEEEoverridecommandlockouts                              

\usepackage{mathptmx} 
\usepackage{times} 
\usepackage{amsmath} 
\usepackage{amssymb}  
\usepackage{graphicx}
\usepackage{mathtools}
\usepackage{upgreek}
\usepackage{gensymb}
\usepackage[noadjust]{cite}
\usepackage{placeins}
\usepackage{xcolor}
\usepackage{url}
\usepackage{mwe}
\usepackage{balance}
\usepackage{multirow}

\DeclarePairedDelimiter\abs{\lvert}{\rvert}%
\DeclarePairedDelimiter\norm{\lVert}{\rVert}%

\makeatletter
\let\oldabs\abs
\def\abs{\@ifstar{\oldabs}{\oldabs*}}
\let\oldnorm\norm
\def\norm{\@ifstar{\oldnorm}{\oldnorm*}}
\makeatother

\title{\LARGE \bf
Mobility Analysis of Screw-Based Locomotion and Propulsion in Various Media
}

\author{Jason Lim$^{\dagger,1}$, Calvin Joyce$^{\dagger,2}$, Elizabeth Peiros$^1$, Mingwei Yeoh$^1$, Peter V. Gavrilov$^2$, Sara G. Wickenhiser$^2$,\\ 
Dimitri A. Schreiber$^1$, Florian Richter$^1$, Michael C. Yip$^1$ \IEEEmembership{Senior Member, IEEE}
\thanks{$^\dagger$ Equal Contribution.}
\thanks{$^1$ J. Lim, E. Peiros, M. Yeoh, D.A. Schreiber, F. Richter, and M.C. Yip are with the Electrical and Computer Engineering Department, University of California San Diego, La Jolla, CA 92093 USA. {\tt\footnotesize\{jkl009, epeiros, myeoh, dschreib, frichter, yip\}@ucsd.edu}}%
\thanks{$^2$ C. Joyce, P.V. Gavrilov, and S. Wickenhiser are with the Mechanical and Aerospace Engineering Department, University of California San Diego, La Jolla, CA 92093 USA. {\tt\footnotesize \{cajoyce, pgavrilo, swickenhiser\}@ucsd.edu}}%
}

\begin{document}

\maketitle
\thispagestyle{empty}
\pagestyle{empty}


\begin{abstract}
Robots ``in-the-wild" encounter and must traverse widely varying terrain, ranging from solid ground to granular materials like sand to full liquids. Numerous approaches exist, including wheeled and legged robots, each excelling in specific domains. Screw-based locomotion is a promising approach for multi-domain mobility, leveraged in exploratory robotic designs, including amphibious vehicles and snake robotics. However, unlike other forms of locomotion, there is limited exploration of the models, parameter effects, and efficiency for multi-terrain Archimedes screw locomotion. 
In this work, we present work towards this missing component in understanding screw-based locomotion: comprehensive experimental results and performance analysis across different media.
We designed a mobile test bed for indoor and outdoor experimentation to collect this data.
Beyond quantitatively showing the multi-domain mobility of screw-based locomotion, we envision future researchers and engineers using the presented results to design effective screw-based locomotion systems.
\end{abstract}

\section{Introduction}

One challenge facing mobile robotics is finding a method of locomotion that can efficiently navigate a wide range of terrains. Wheeled or tracked vehicles are great for traversing solid ground \cite{wong_2001}, legged locomotion has shown to be adept at climbing obstacles \cite{Roennau_2014, guizzo_2019}, and many designs exist for swimming robots \cite{chu_2012}. However, these methods fail quickly outside the specific environments or media for which they were developed. An example of traditional locomotion methods' shortcomings are NASA's Mars rovers that have gotten stuck in soft sand \cite{li_2008}. One proposed method that shows great potential in navigating many different types of media is screw-propulsion \cite{arcsnake_icra}. The ARCsnake robot, which combines screw propulsion with a snake-like backbone, shows the concept for a robotic platform proposal for NASA's mission to search for extant life in the subterranean ocean of Saturn's Moon, Enceladus \cite{carpenter2021exobiology}, due to its multi-domain mobility capabilities \cite{arcsnake_tro}.

Archimedes screws were originally invented for transporting water and continue to be widely used as pumps for all types of fluids and granular media. Over the years, they have also been used as drilling mechanisms, injection molding devices, and turbines \cite{Waters_2015}. The first applications of screws were as propellers were for watercraft \cite{rossell_chapman_1962, wells_1841}, and the realization that screws could generate propulsive forces on land led to them being proposed for amphibious vehicles \cite{neumeyer_1965}. By the mid-20th century, several screw-propelled vehicles had been developed that demonstrated locomotion and towing capabilities in many types of media, including soil, marshes, snow, and ice \cite{Freeberg_2010}. Screw locomotion naturally found its way from large-scale vehicles to smaller mobile robotics \cite{arcsnake_icra, arcsnake_tro, Nagaoka_2009, lugo_2016, osinski_2015, liang_2011}. Screw locomotion has even been studied for medical applications as a method for traversing the mix of solid and liquid media in the intestines \cite{Mayfield_2015, kim_2010}. The multi-domain viability of screw propulsion provides many advantages for exploratory mobile robots in situations where traditional methods of locomotion fail. It is still not well understood how screw-locomotion performance metrics, such as generated thrust, degree of slip, and traveling speed, change across different terrains, and how well current models can predict these metrics.


\begin{figure}
    \centering
    \vspace{2mm}
    
    \includegraphics[trim={0 5.55cm 0 0},clip, width=\linewidth]{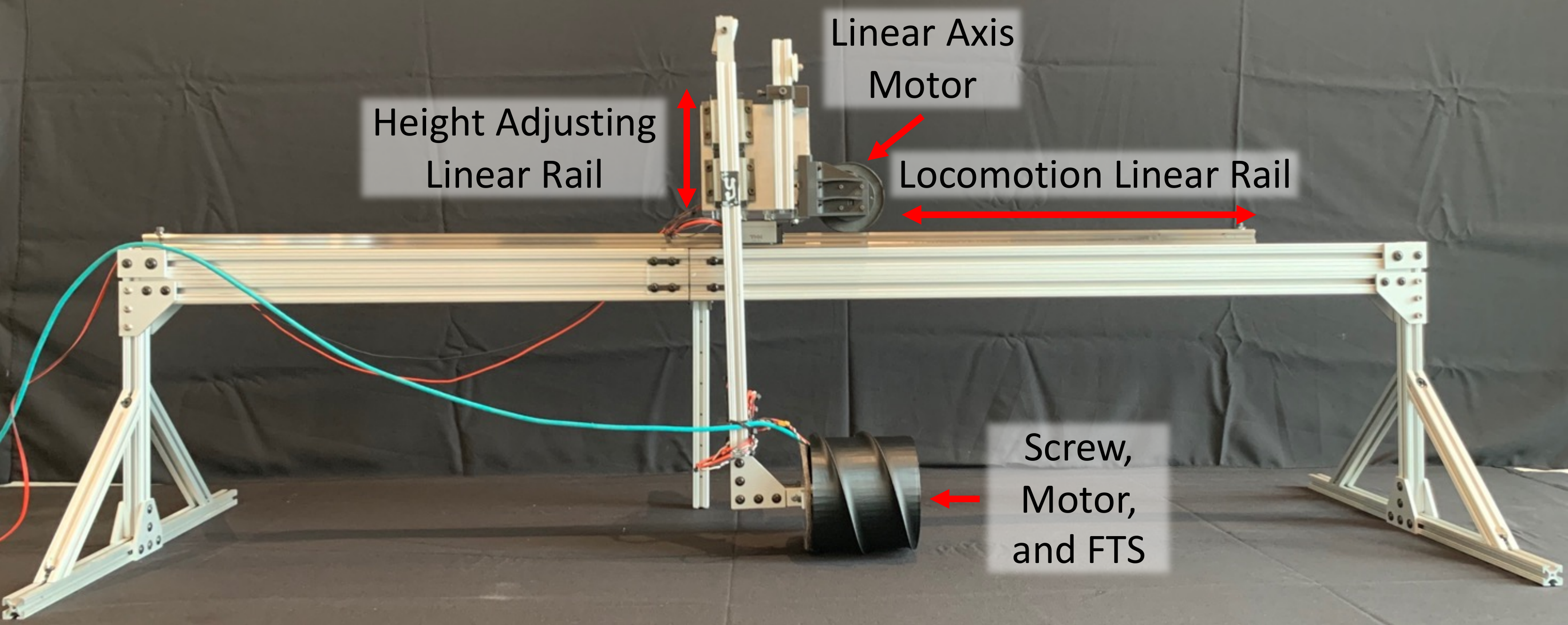}
    
    \vspace{1mm}
    
    \includegraphics[width=0.49\linewidth]{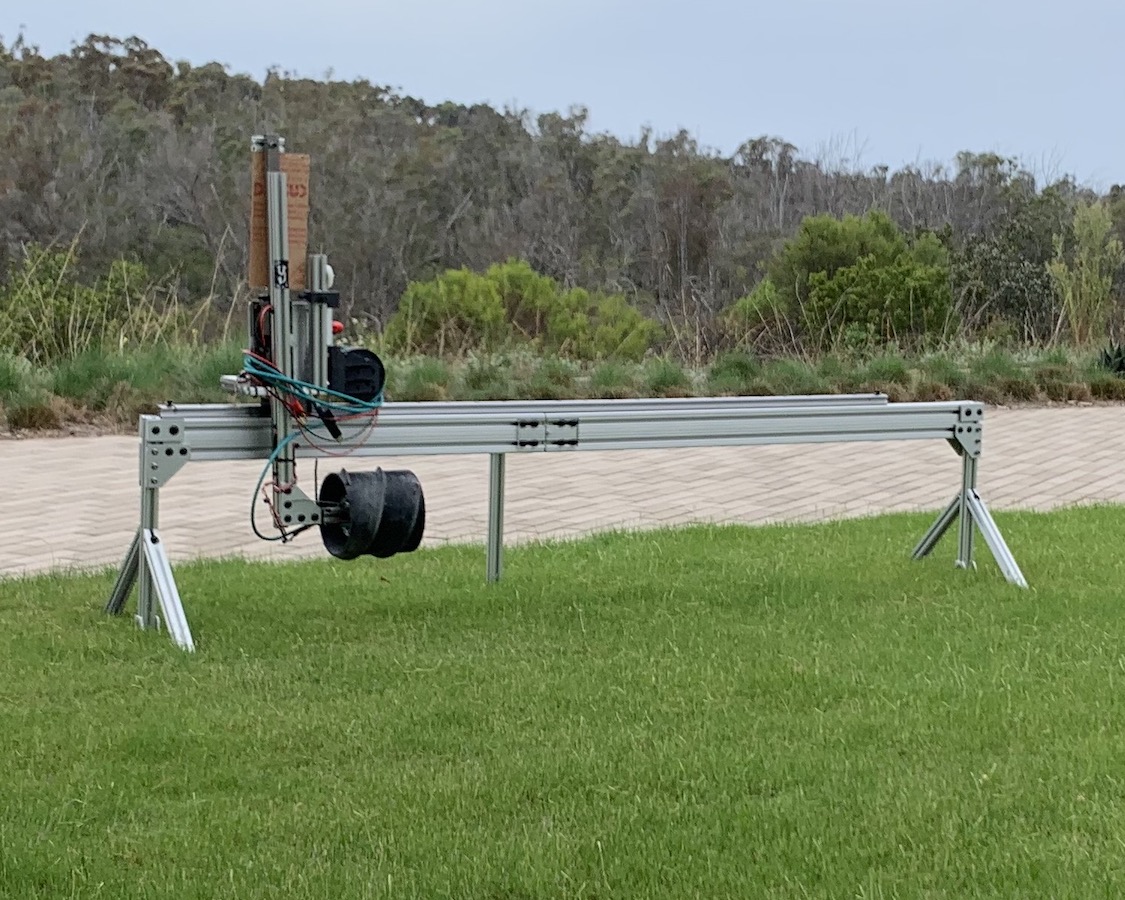}
    \includegraphics[width=0.49\linewidth]{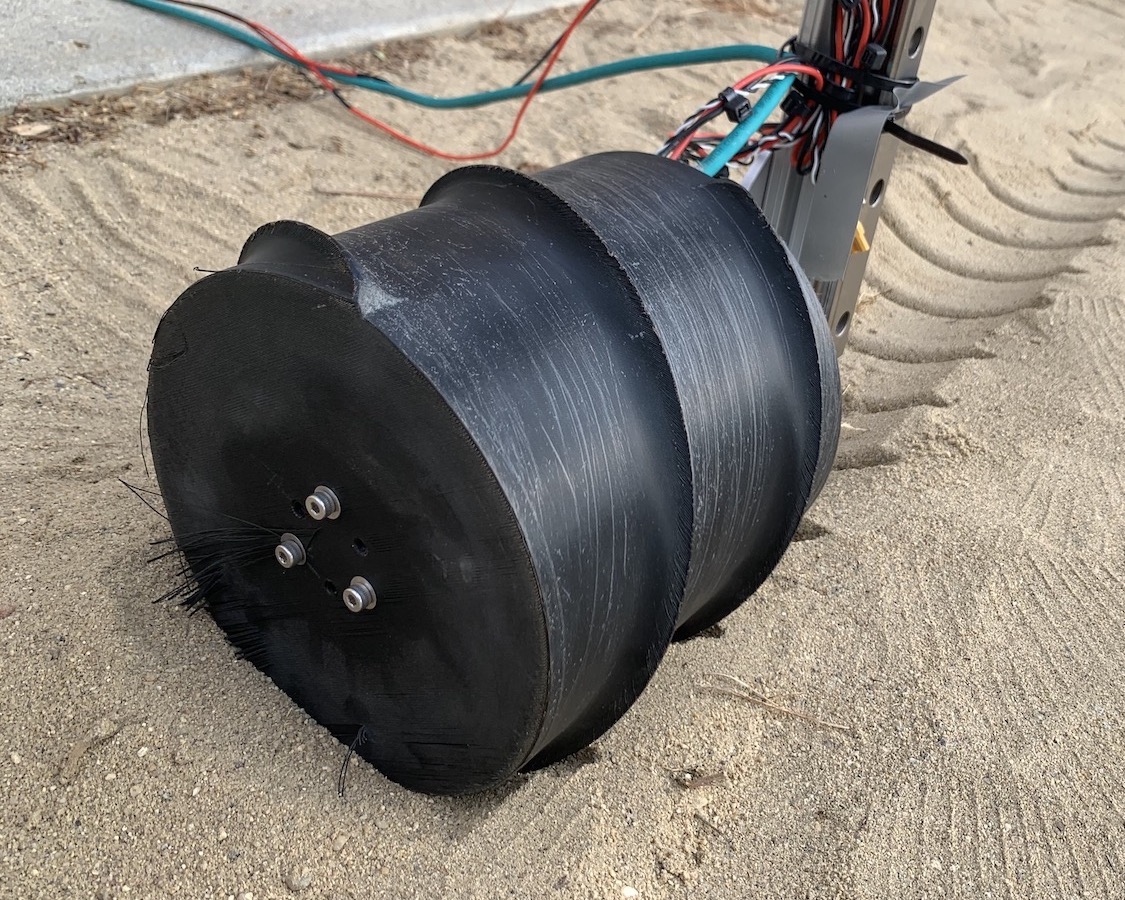}
    \caption{Mobile test bed for collecting performance of screws for locomotion.
    The top figure shows a diagram of all the test bed's major components.
    The resulting velocity and thrust force are read with a Linear Axis Motor and 6-DoF Force Torque Sensor respectively.
    The Linear Axis Motor can also be used to apply an axial load.
    The mobility of the test bed enables easy deployment in real-world environments such as grass and sand (bottom row).
    }
    \label{fig:system_overview}
    
    \vspace{-5mm}
\end{figure}

\subsection{Related Works}

Screws can be thought of as devices that convert rotational force into linear force due to the helical shape of their threads. The physics of screw interaction with a typical threaded nut has been well studied \cite{budynas2011shigley}; used in this way, the screw is fully engulfed such that all forces cancel out except in the direction of the screw axis. When screws are used for propelling vehicles or robots, they often sit on top of or partially submerged in media and the net force of a single screw cannot be purely along the axis. Multiple screws are needed to generate forces in multiple directions, and thus the first screw vehicle designs consisted of two parallel counter-rotating screws \cite{neumeyer_1965}. 

After being empirically shown that screw propulsion is useful for multi-domain mobility, several researchers attempted to model the interaction of screws and deformable media such as sand and water \cite{Cole_1961, dugoff_1967}. These first models relied on fairly bold assumptions, for example ignoring the energy imparted to the displacement of media and assuming no slip conditions. Other groups have more recently investigated the locomotion characteristics of parallel screw rovers, using simplified kinematic models to simulate locomotion capabilities \cite{He_2017, lugo_2016, osinski_2015}. To forgo some of the assumptions that may lead to an inaccurate model, a model based on terramechanics principles has been proposed to capture the relationship of slip and locomotion and account for the energy lost to displacing the media \cite{Nagaoka_2010}. Screw locomotion models have been experimentally confirmed in a few media such as sand, ice, and water \cite{Cole_1961, Nagaoka_2010, dugoff_1967}. Additionally, screw locomotion performance has been demonstrated qualitatively but not quantitatively across an even wider variety of terrains \cite{Freeberg_2010, oshima_1982}.
Overall, previous experimental research on screw locomotion has been fairly limited in terms of the metrics used to analyze performance. To connect our results with traditional off-road vehicle performance evaluation methods we show a comprehensive set of performance metrics, including novel metrics for screw-based locomotion such as cost of transport.

\subsection{Contributions}


%

While there is sufficient evidence that screw locomotion is effective for multi-domain mobility, quantitative analysis of locomotion performance across a large variety of domains has not been presented before.
Without robust experimental performance metrics, mechanical and control design for screw-based systems are unable to be optimized for their specific use cases.
To this end, we present the following contributions:
\begin{enumerate}
    \item experimentation for screw-based locomotion performance across a large variety of medium,
    \item novel analysis of screw-based locomotion capabilities highlighting efficiency and cost of transport, and
    \item a mobile test bed design able to test the locomotion capabilities of single and multi-screw configurations in both lab and field settings.
\end{enumerate}
Through our comprehensive real-world results, we validate that screw-based locomotion is effective in a large array of media.
We envision that future researchers and engineers will use the presented empirical results to design effective screw-based locomotion systems.

\section{Mobile Test Bed Design}

The mobile test bed is designed to collect then locomotion performance of screws in different media. An overview of its design is shown in Fig. \ref{fig:system_overview}.
To ensure high-quality measurements across a large range of media, the test bed must have the following requirements:
\begin{itemize}
    \item Constrain the motion to a single, linear axis to isolate the locomotion measurement.
    \item Lightweight and easy to reposition for both indoor and outdoor experiments.
    \item Apply axial loads to simulate towing.
    \item Sufficient stiffness such that deflection does not impact experimental data.
    \item Measure the screw's velocity and applied torque.
    \item Measure the resulting screw-locomotion's velocity and applied forces and torque.
\end{itemize}
The coming sub-sections provide design details to meet the listed system requirements and an experimental procedure for collecting data with the mobile test bed. 



\begin{figure}[b]
    \centering
    \vspace{-5mm}
    \includegraphics[width=\linewidth]{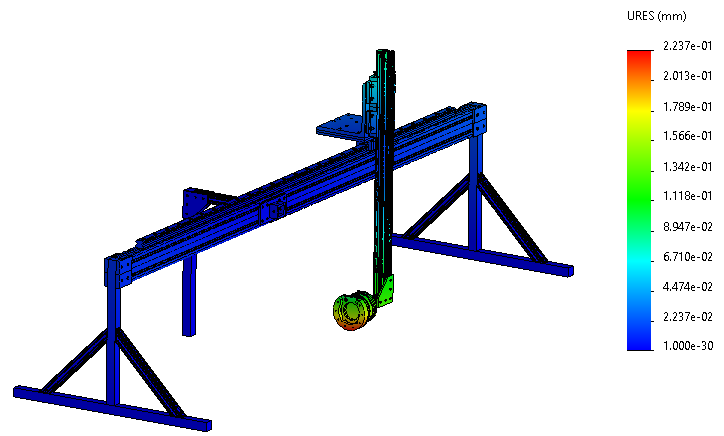}
    \caption{FEA on mobile test bed where the 5 Nm is applied on where the screw would be located. The direction and magnitude of the applied force in this analysis is a worst-case scenario and sufficiently larger than what will be seen in the experiments.
    }
    \label{fig:test_bed_fea}
\end{figure}

\subsection{Mechanical Design}

To constrain the locomotion to a single, linear axis and keep the structure lightweight, a linear rail 	Ball Bearing Carriage for 23 mm Wide Rail 6709K16 with 1600 mm length is mounted on ``T-Slotted Framing Double Six Slot Rail, Silver, 3" High x 1-1/2" Wide, Hollow" (47065T108).
Similarly, the legs of the test bed are made with ``T-Slotted Framing Silver Diagonal Brace for 1" High Single Rail, 12" Long" (47065T188).
The screw(s) for locomotion is attached to the linear rail's carriage Ball Bearing Carriage with Flange for 23 mm Wide Rail 6709K15 and another linear rail, 23 mm Wide Guide Rail for Ball Bearing Carriage 6709K53, for height adjustment.
The height-adjusting linear rail helps compensates for potentially uneven surfaces being experimented on.
An additional motor, RMD-L 7015, is placed on the carriage of the linear rail and can be held passive or regulate a torque hence applying an axial load on the system.
The RMD-L 7015 is selected due to its high transparency since it is brush-less and does not have an internal gearbox.
We decided on a screw with a flat face since it is representative of previous screw robot designs \cite{arcsnake_icra}.
Finally, the screw mounted for experimentation can be easily re-configured.
In our experiments, we used single and parallel screw configurations.
All the mechanical design choices are validated for minimal deflection with a Finite Element Analysis (FEA) as shown in Fig. \ref{fig:test_bed_fea}.
The maximum deflection from the FEA is less than 0.3 mm which is sufficiently low for locomotion experiments.

\subsection{Sensors for Measurements}

The screw being experimented on is driven with an RMDx8 Pro motor which provides velocity measurements from a quadrature optical encoder to measure the screw's angular velocity.
The linear axis motor, RMD-L 7015, is used to measure the screw-locomotion's velocity.
Finally, a 6-DoF Force Torque Sensor (FTS), the Axia80 (ATI Industrial Automation), is positioned near the center of mass of the driving screw to measure the screw's applied torque and the resulting in screw-locomotion forces.

\subsection{Experimental Setup}

To ensure consistent results, the following experimental setup steps are taken after placing the mobile test bed:
\begin{enumerate}
    \item Flatten and level the media to achieve as uniform conditions as possible.
    \item The height-adjustable linear rail is locked such that the screw(s) are free hanging. A ``free hanging" measurement from the FTS is taken to capture any potential drifts between trials.
    \item The height-adjustable linear rail is unlocked and the screw is set down on the media. A ``set down" measurement from the FTS is taken to measure any pre-loading from the media on the screw.
    \item The FTS sensor is zeroed to then measure differential measurements. 
\end{enumerate}
From this point, the screw motor can be driven to begin the experiment.



\section{Performance Evaluation}

\begin{figure*}
    \centering
    \vspace{2mm}
    \includegraphics[width=0.136\linewidth]{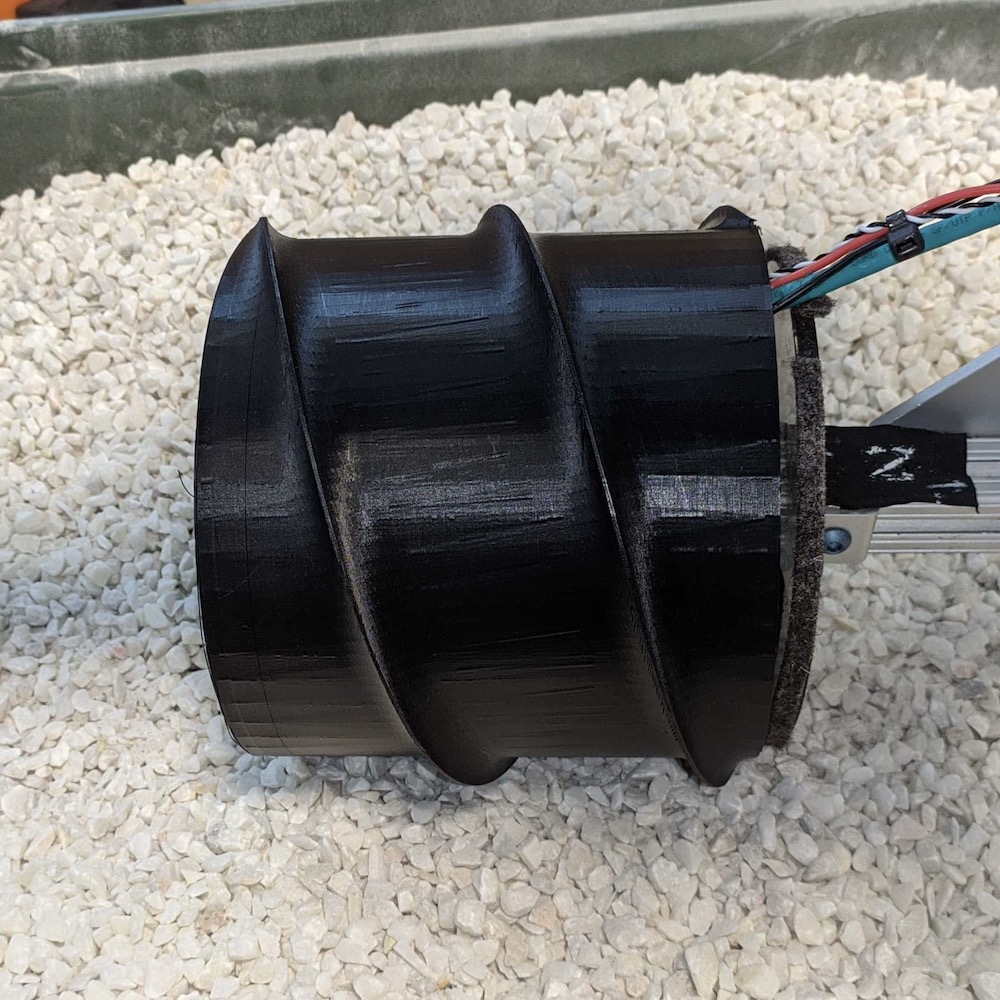}
    \includegraphics[width=0.136\linewidth]{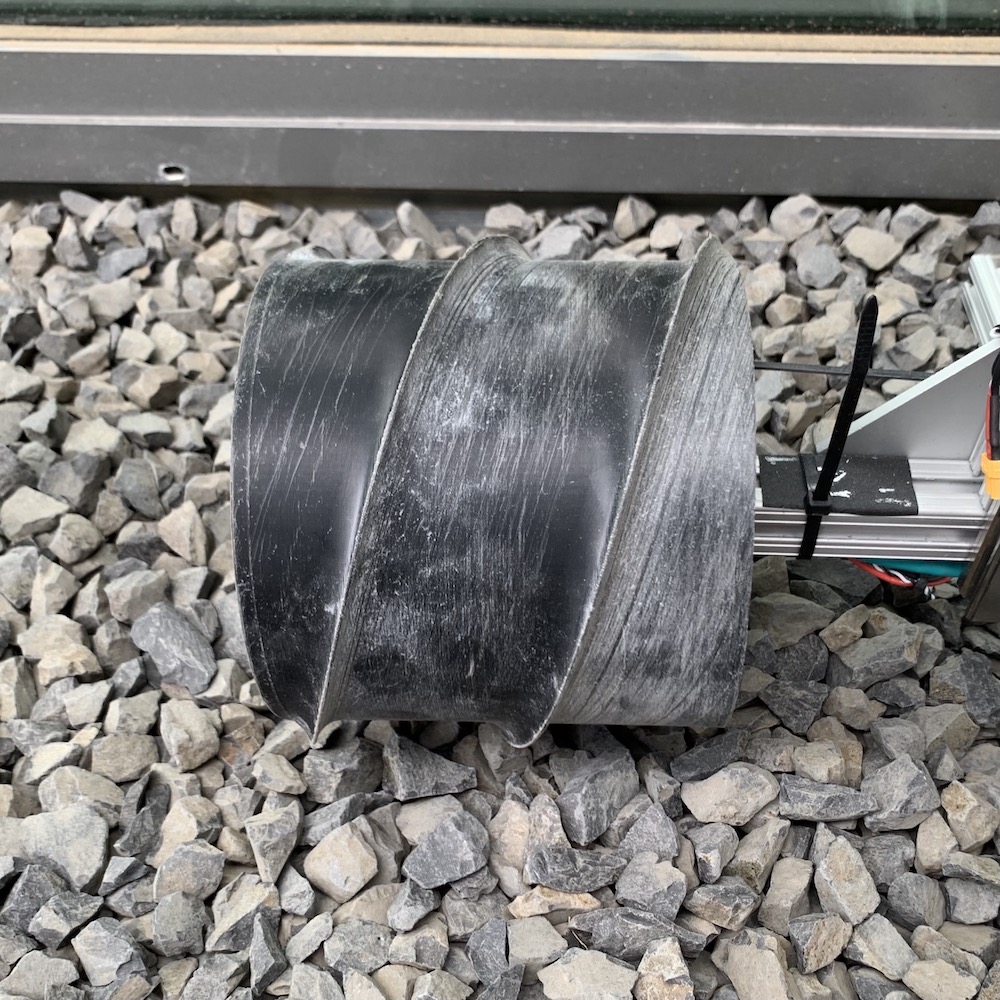}
    \includegraphics[width=0.136\linewidth]{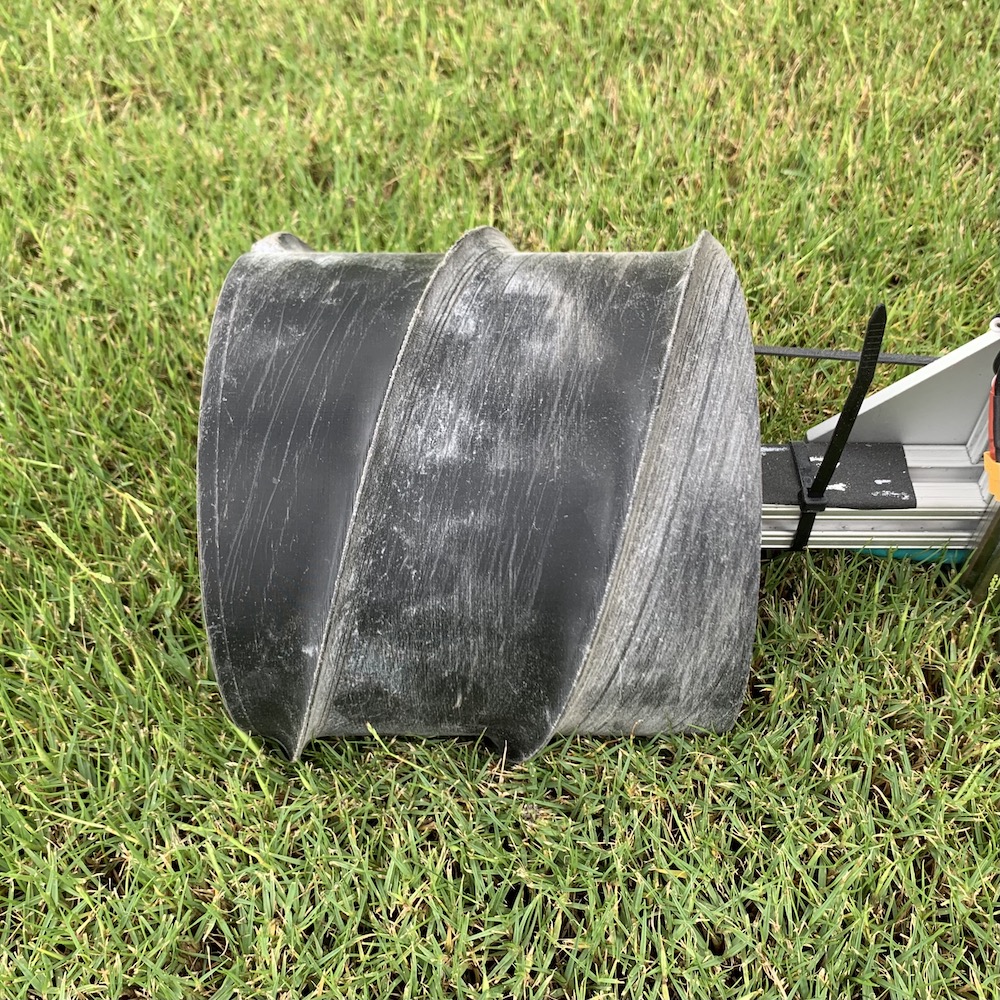}
    \includegraphics[width=0.136\linewidth]{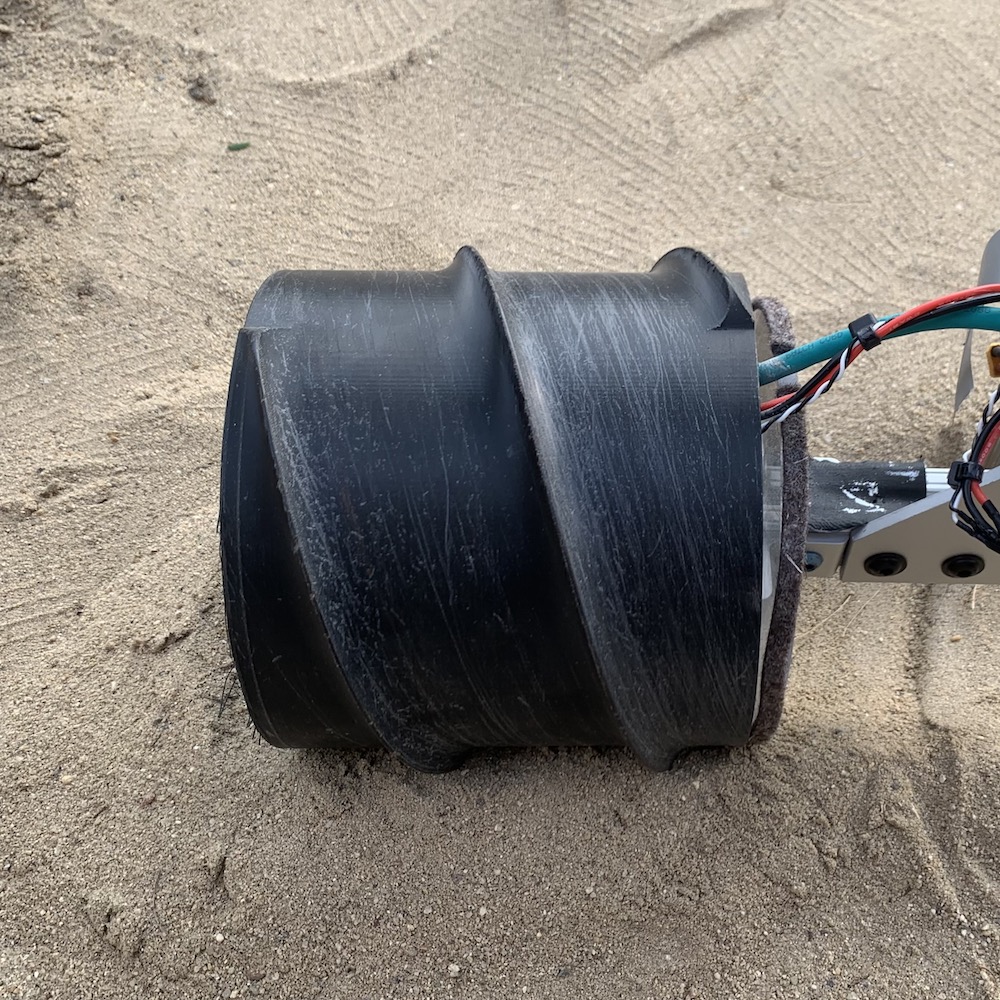}
    \includegraphics[width=0.136\linewidth]{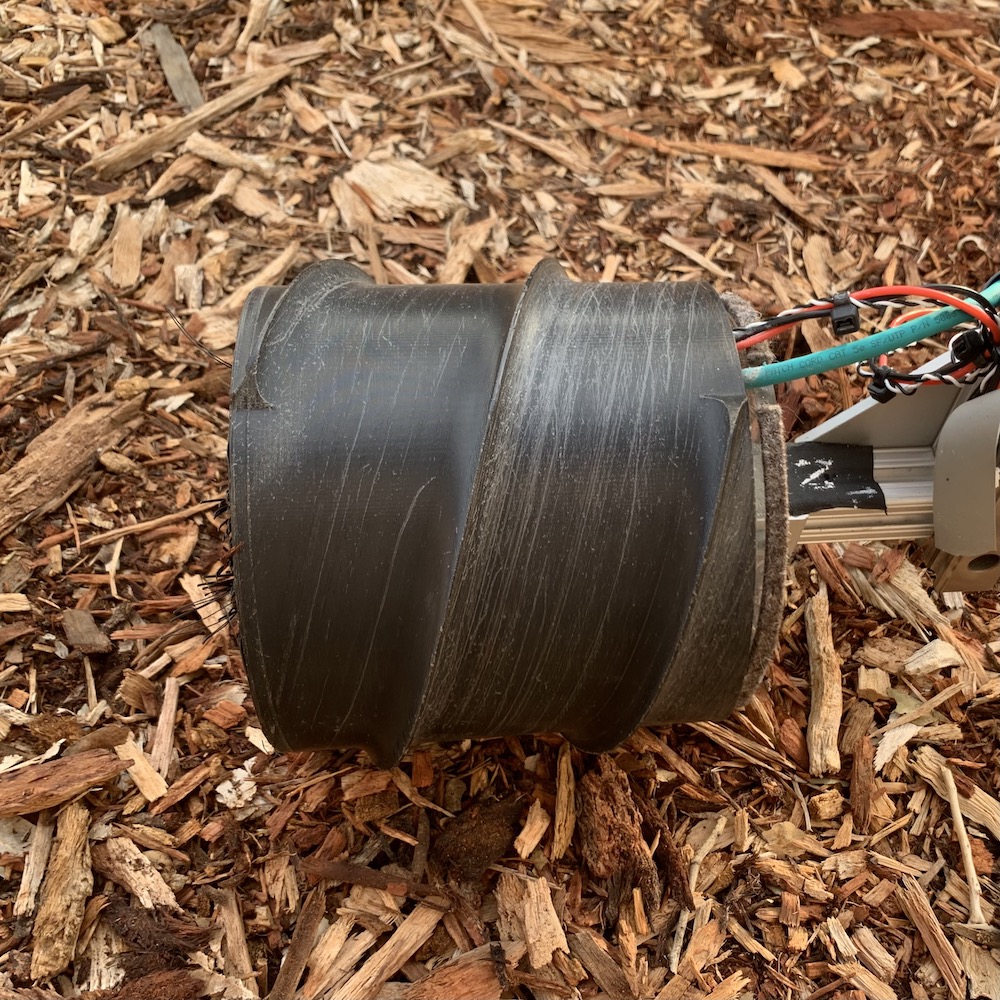}
    \includegraphics[width=0.136\linewidth]{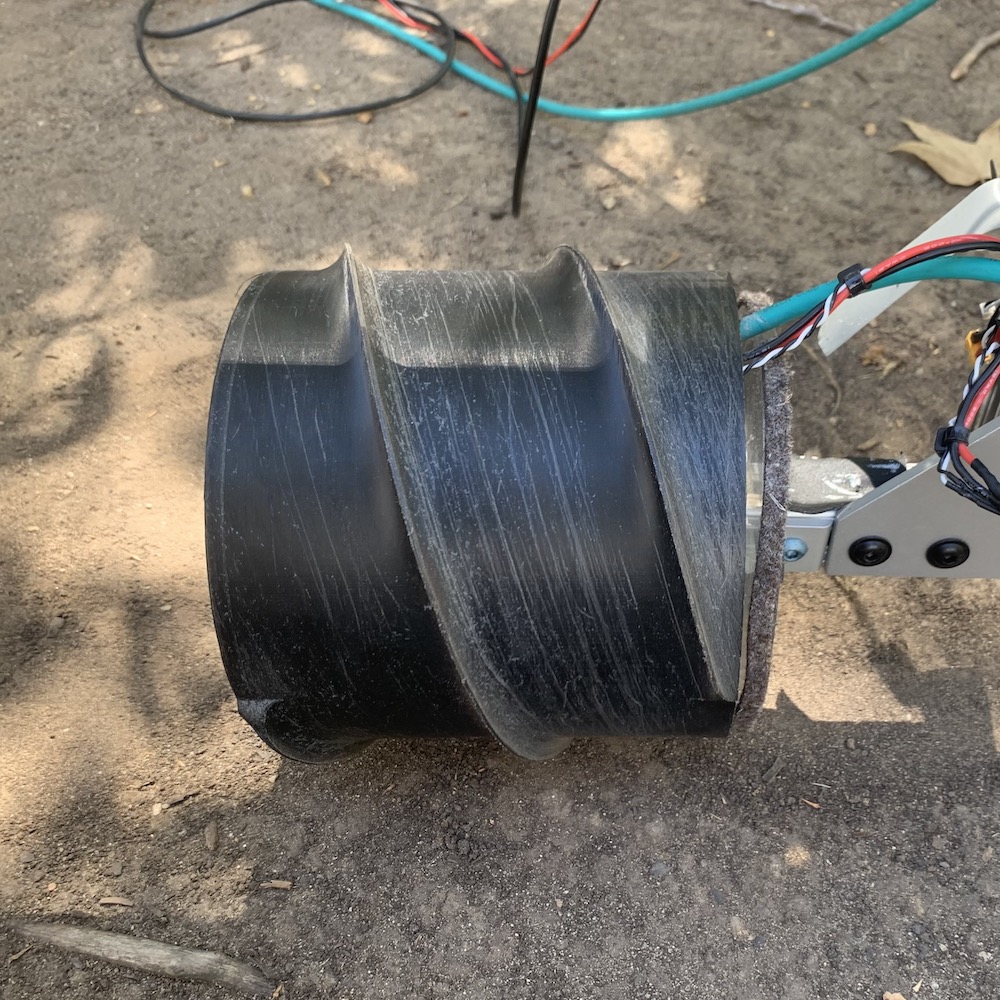}
    \includegraphics[width=0.136\linewidth]{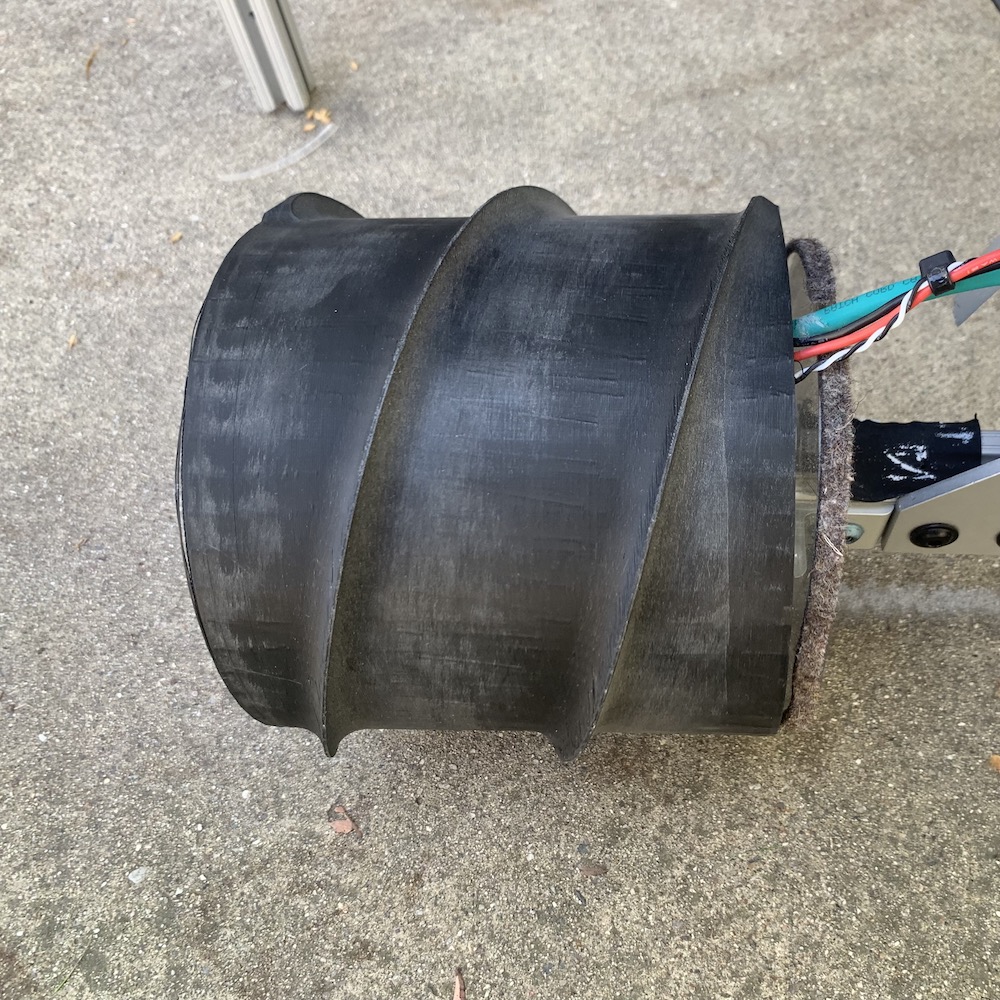}
    \caption{Representative images of the test bed experiments conducted in the following media (from left to right): small gravel, big gravel, grass, sand, wood chips, dirt, and concrete.}
    \label{fig:collage_from_experiments}
    \vspace{-1mm}
\end{figure*}

\begin{figure*}[t]
    \centering
    \includegraphics[trim={1cm 1.5cm 2.5cm 3.5cm},clip,width=0.49\linewidth]{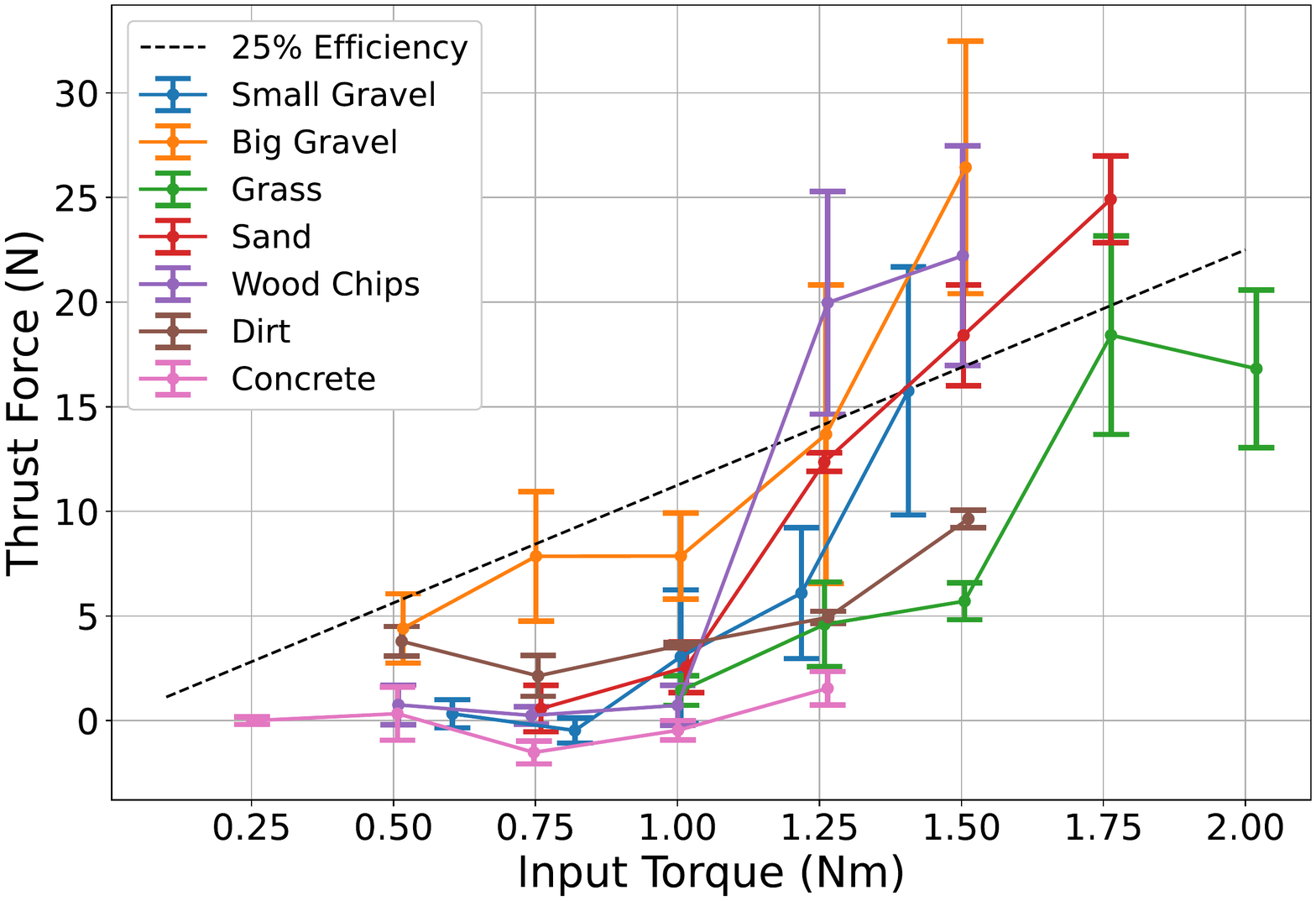}
    ~
    \includegraphics[trim={1cm 1.5cm 2.5cm 3.5cm},clip,width=0.49\linewidth]{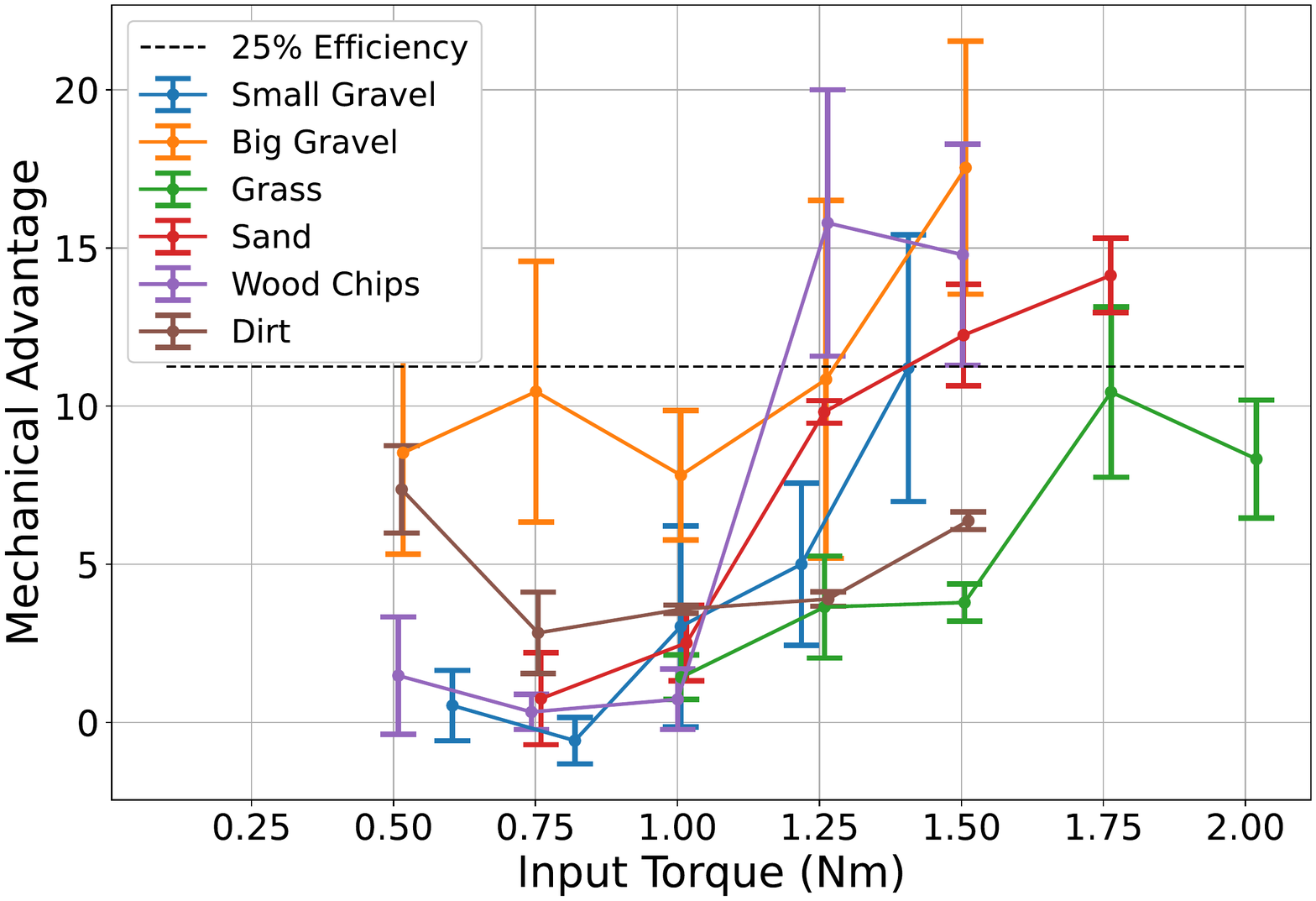}
    \caption{The left and right plots show the Thrust Force and MA, respectively, for different input torques in a static condition. The max torque tested on is the maximum torque the screw is held static for the media.}
    \label{fig:static_results}
\end{figure*}

\begin{figure*}[t]
    \centering
    \includegraphics[trim={1cm 1.5cm 2.5cm 3.5cm},clip, width=0.48\linewidth]{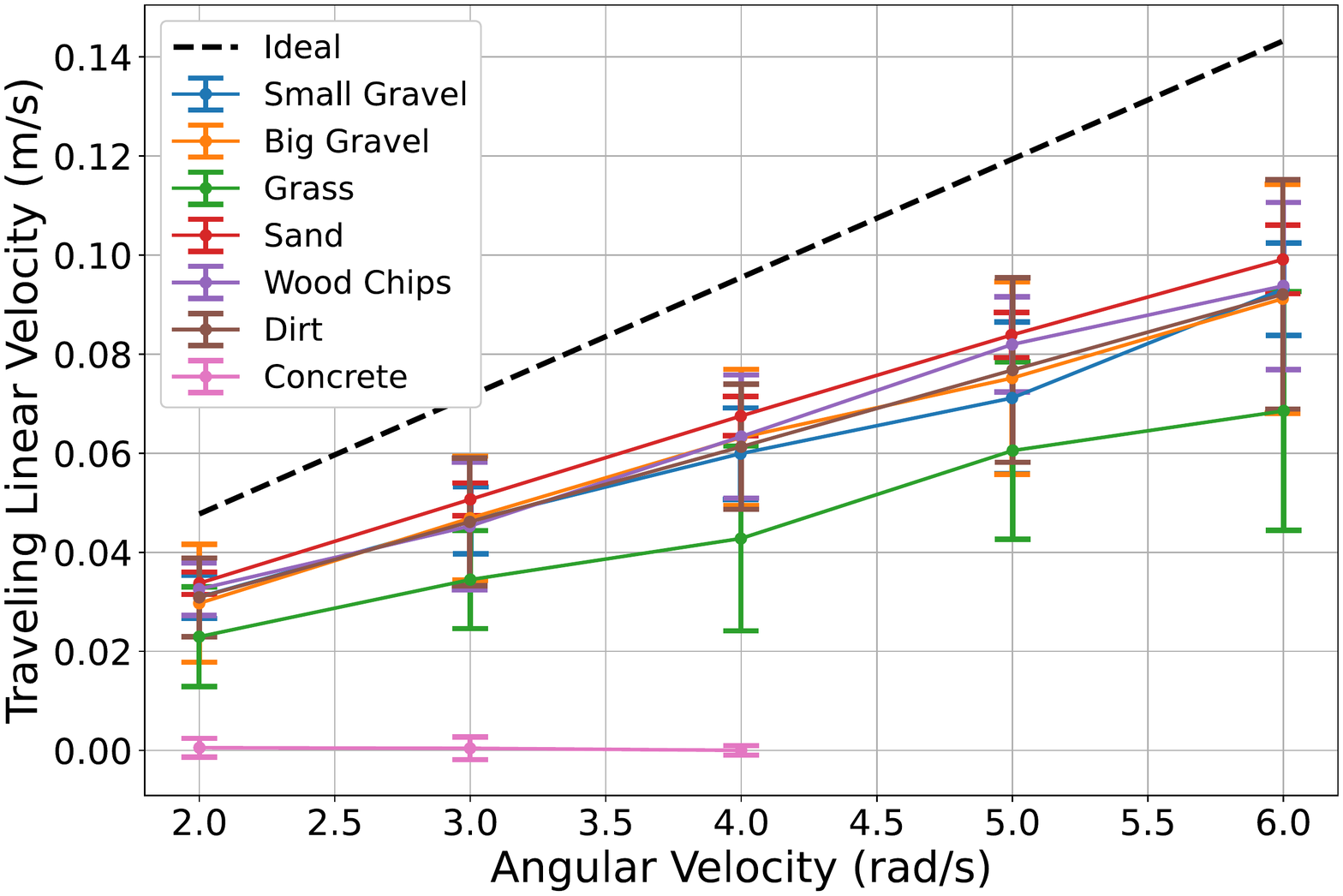}
    ~~~
    \includegraphics[trim={1cm 1.5cm 2.5cm 3.5cm},clip, width=0.48\linewidth]{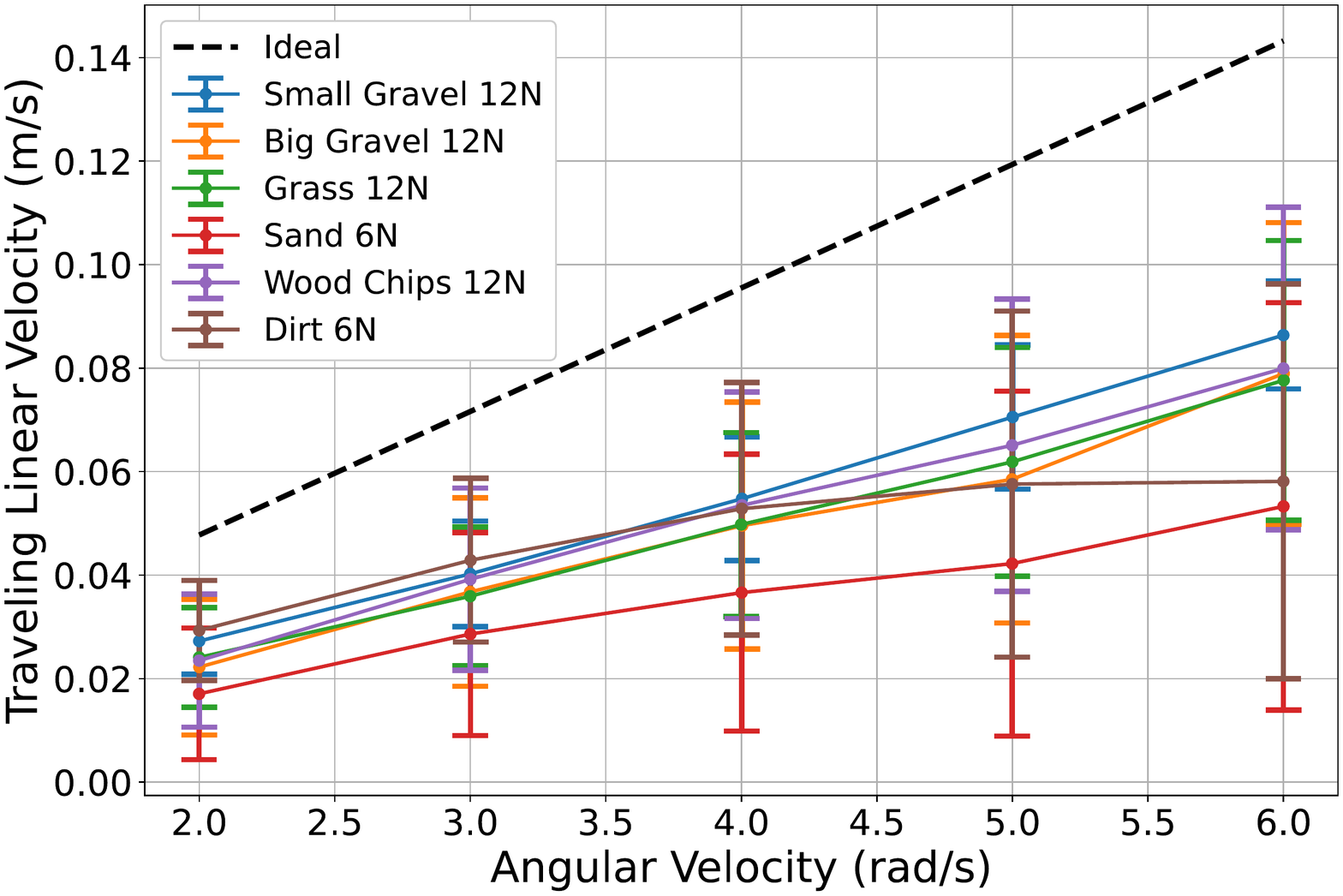}
    \caption{The left and right plots show the linear traveling velocity as a function of input angular velocity with and without an axial load, respectively. As visible from these plots, the linearity of the velocity relationship holds across all media except concrete which never produced any motion. }
    \label{fig:single_vel}
    \vspace{-2mm}
\end{figure*}

The screw performance is evaluated in static and motion scenarios.
The test bed supports both of these metrics by simply locking/unlocking the locomotion linear rail. 
The metrics are inspired by both screw and conventional locomotion analysis hence providing good coverage of metrics to understand screw-locomotion across different media.

\subsection{Static Performance}

For typical screws in a threaded environment, the relationship between the input torque and the amount of force generated in the axial direction can be described by the mechanical advantage (MA). Derived from the conservation of energy principles, mechanical advantage is:
\begin{equation}
    MA = \frac{F_{thrust}}{\tau_{in}} = \eta_s \frac{2\pi}{\ell},
    \label{eq:ma}
\end{equation}
where $F_{thrust}$ is the thrust force measured by the FTS, $\tau_{in}$ is the torque applied by the screw motor, $\eta_s$ is the static efficiency and $\ell$ is the longitudinal distance a thread spans in one full revolution around the screw (i.e. pitch times the number of thread starts). Mechanical advantage is heavily dependent upon the frictional properties of the media. For media that is not perfectly rigid, additional efficiency losses may result from deformations of the media, and therefore $\eta_s$ can be treated as the term that encompasses any type of energy loss to surroundings. Comparing the mechanical advantage and efficiency of the screw across different media provides insight into the performance of the screw in a static scenario. 

\subsection{Motion Performance}

The ideal locomotion velocity is given by assuming that the screw threads move through grooves as if it were in a threaded nut. Therefore, the ideal velocity for a screw can be computed geometrically as follows:
\begin{equation}
    v_{ideal} = \frac{\omega \ell}{2\pi}
\end{equation}
where $\omega$ is the angular velocity of the screw motor.
From the ideal velocity, a longitudinal slippage can be defined as:
\begin{equation}
    s = \frac{ v_{ideal} - v}{v_{ideal}},
\end{equation}
where $v$ is the measured velocity from the screw-locomotion.
The slip factor equals 1 when the screw slips uncontrollably and doesn't move at all, and equals 0 when the screw moves at the ideal optimal velocity hence giving a good metric about motion capabilities of screws in different media. 

From the measured traveling velocity, the locomotive efficiency of the system, $\eta_m$, is calculated as
\begin{equation}
    \eta_m = \frac{ F_{thrust} v}{\tau_{in} \, \omega} 
\end{equation}
and the cost of transport, $COT$, is calculated as
\begin{equation}
    COT = \frac{P_{in}}{m g v},
\end{equation}
where $P_{in}$ is the power input to the system, $m$ is mass (i.e. screw(s), motors, and vertical rail), and $g = 9.81$ $m/s^2$.
Locomotion efficiency is different from static efficiency, $\eta_s$, since it incorporates motion.
Cost of transport (COT) is a dimensionless quantity related to energy efficiency that allows for comparison among different modes of locomotion.

\section{Experiments and Results}


Experiments are conducted using the mobile test bed on a wide range of types of media in both a lab setting and in real world environments as shown in Fig. \ref{fig:collage_from_experiments}.
These experiments characterize the locomotion performance by measuring the velocity and output forces of the screw in these different media.
We also experiment with a parallel screw configuration to show how the single screw measurements can be used to estimate the performance of previously developed screw-based locomotion systems \cite{neumeyer_1965, He_2017}.


\subsection{Implementation Details}
Experiments are conducted with a 3-D printed ABS screw of root radius 77.5 mm, outer radius 86 mm, lead angle 16\degree, the total length of 147 mm, and two thread starts.
These screw parameters are selected since they have been shown to be most efficient in generating thrust force \cite{Cole_1961} and have been the standard parameters for recent screw-based systems \cite{arcsnake_tro, Freeberg_2010}.
To process the raw data from the test bed for analysis, the data from the motors and FTS are passed through a low pass filter with a cutoff frequency of 6 Hz and a sampling frequency of 125 Hz.
The data is clipped manually to only include the steady state portion of the experiment.

\begin{table*}[t]
    \centering
    \caption{Results from linear motion experiments with and without an axial load applied.}
    \resizebox{\textwidth}{!}{
    \begin{tabular}{c|c|c|c|c|c|c|c|c|c}
    \hline
        \multirow{2}{*}{Media} & Coefficient & \multicolumn{2}{c|}{Longitudinal  Slippage} & \multicolumn{2}{c|}{Thrust in Motion} & \multicolumn{2}{c|}{Locomotive Efficiency} & \multicolumn{2}{c}{COT}\\ 
        & of Friction & \textit{No  Load} & \textit{w/ Load} & \textit{No Load} & \textit{w/ Load} &  \textit{No Load} & \textit{w/ Load} & \textit{No Load }& \textit{w/ Load} \\
        \hline \hline
        Small Gravel & $0.32 \pm 0.01$ & $0.37 \pm 0.10$ & $0.42 \pm 0.12$ & $5.63 \pm 0.9$ & $16.9 \pm 1.2$ & $0.059$ & $0.17$ & $2.44$ & $2.53$\\
        Big Gravel & $0.23 \pm 0.03$ & $0.36 \pm 0.18$ & $0.49 \pm 0.24$ & $6.99 \pm 1.9$ & $18.0 \pm 2.8$ & $0.093$ & $0.17$ & $1.91$ & $2.79$ \\
        Grass & $0.38 \pm 0.05$ & $0.52 \pm 0.17$ & $0.48 \pm 0.19$ & $7.32 \pm 1.2$ & $18.5 \pm 1.8$ & $0.046$ & $0.17$ & $4.03$ & $2.84$ \\
        Sand & $0.48 \pm 0.03$ & $0.30 \pm 0.04$ & $0.63 \pm 0.27$ & $8.53 \pm 1.0$ & $9.67 \pm 1.6$ & $0.082$ & $0.05$ & $2.63$ & $5.24$\\
        Wood Chips & $0.31 \pm 0.04$ & $0.34 \pm 0.12$ & $0.46 \pm 0.24$ & $7.21 \pm 1.5$ & $18.1 \pm 2.8$ & $0.092$ & $0.20$ & $1.99$ & $2.30$\\
        Dirt & $0.38 \pm 0.02$ & $0.36 \pm 0.16$ & $0.47 \pm 0.24$ & $9.06 \pm 1.8$ & $11.4 \pm 2.2$ & $0.092$ & $0.099$ & $2.50$ & $3.08$\\
        Concrete & $0.34 \pm 0.03$ & $0.99 \pm 0.02$ & NA &  $2.15 \pm 0.5$ & NA & $0.0003$ & NA & NA & NA\\
        \hline
    \end{tabular}
    }
    \label{tab:media}
\end{table*}

\begin{figure}
    \centering
    \includegraphics[width=0.49\linewidth]{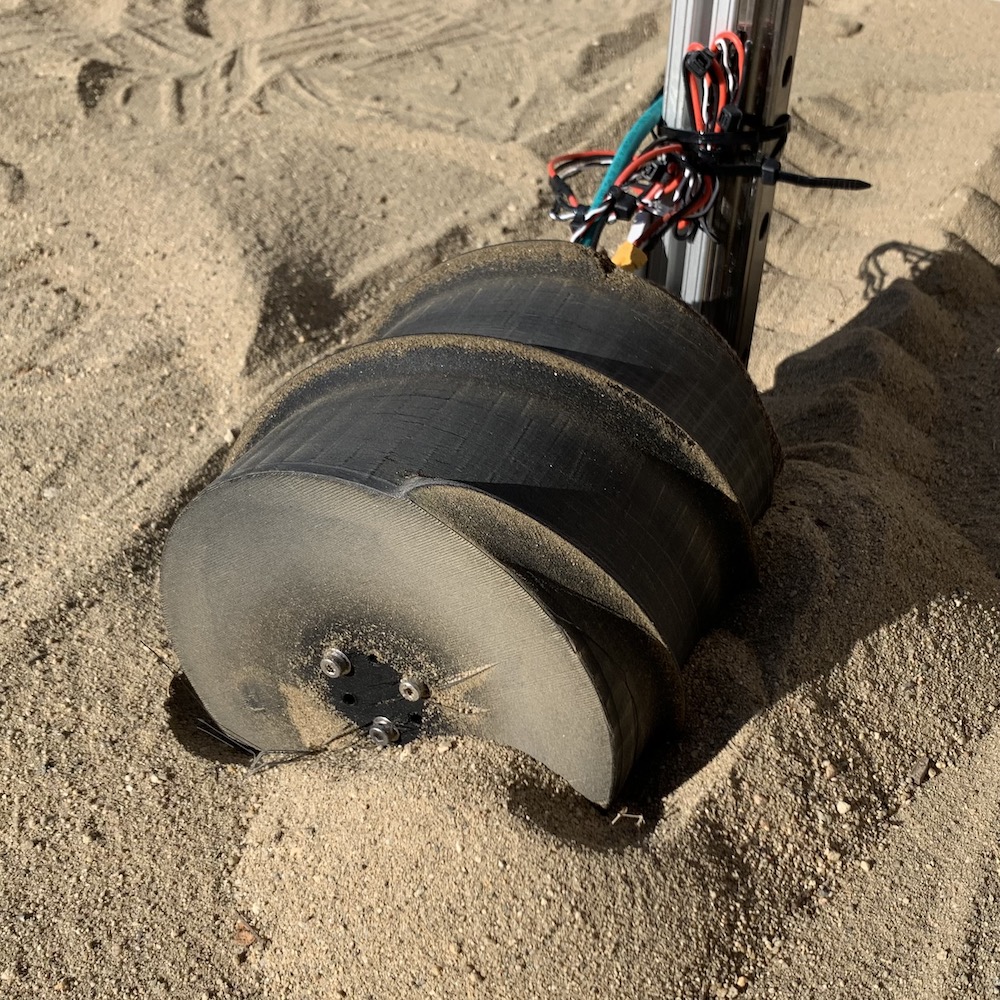}
    \includegraphics[width=0.49\linewidth]{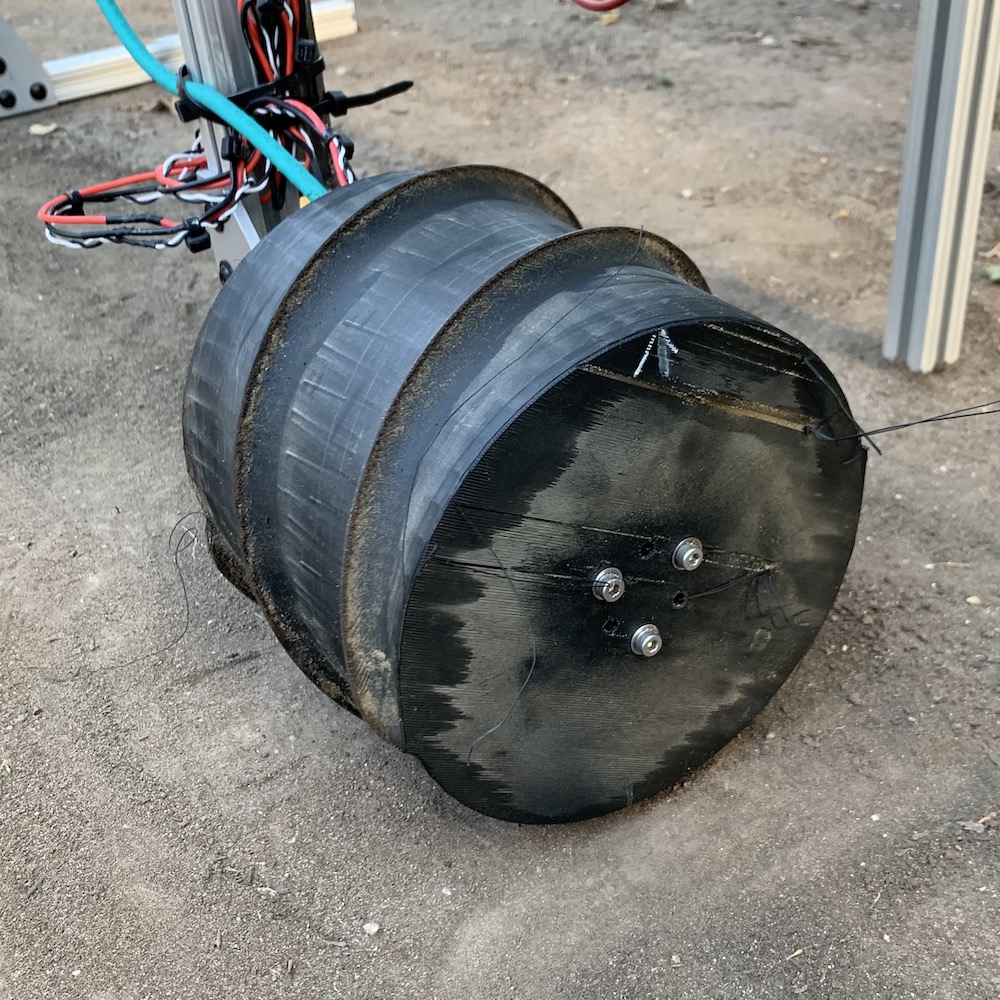}
    \caption{When applying a 12N axial load during the sand and dirt experiments, the screw dug itself into the material.
    For the sand experiment (left), this caused slippage on the test bed's linear axis motor leading to inconsistent measurements.
    In the dirt experiment (right), the screw dug itself into a rock bed hence resulting in the screw slipping and leading to inconsistent results.
    Therefore, for both of these environments, we used a lower axial load of 6N in the linear motion experiments.
    }
    \label{fig:stuck}
    \vspace{-3mm}
\end{figure}

\subsection{Single-Segment}

\subsubsection{Static Performance Test}

For this experiment, the goal is to measure the maximum thrust produced for a given input torque under static conditions.
All degrees of freedom are restricted except for the vertical direction so that the screw is still allowed to sink into the media.
Torque is applied such that the screw may spin initially but eventually gets stuck and a steady state, a static condition exists.
Therefore, the screw motor's max torque for this experiment is set such that the screw remains static in the media being tested.
The ratio of thrust measured during this steady state and the input torque is the MA of the screw and varies depending on the properties of the media.
Results are shown in Fig. \ref{fig:static_results}.

\subsubsection{Linear motion test}

This experiment allows freedom of motion in both the axial and vertical directions, so that the speed of travel can be measured and sinkage into the media is still allowed, while all other degrees of freedom are restricted by the test bed.
The screw motor is set to regulate at a constant velocity: 2, 3, 4, 5, and 6 rad/s.
The measured locomotion velocity, motor torque, angular velocity, and force measurements allow the calculation of metrics such as longitudinal slip, net thrust (also called drawbar pull), locomotive efficiency, and cost of transport.
Furthermore, we repeat the experiments with and without an axial load.
The axial load is set to 12N except for the sand and dirt tests where a load of 6N is used.
A lower axial load is required for sand and dirt because, at 12N, the screw would dig itself in as shown in Fig. \ref{fig:stuck}, leading to inconsistent results.
The results are shown in Fig. \ref{fig:single_vel} and Table \ref{tab:media}.
Note that the concrete results are incomplete due to the lack of mobility the screw provided on that media.

We also measure the coefficient of friction on each media by replacing the screw with a smooth cylinder of the same material. Driving the screw motor at a constant velocity, the coefficient of friction, $\mu_f$, can then be estimated as 
\begin{equation}
    \mu_f = \frac{\tau_s}{m g r},
\end{equation}
where $\tau_s$ is the torque about the screw axis measured by the FTS and $r$ is the radius from the screw axis to the contact point. 

\begin{figure}
    \centering
    \includegraphics[width=0.49\linewidth]{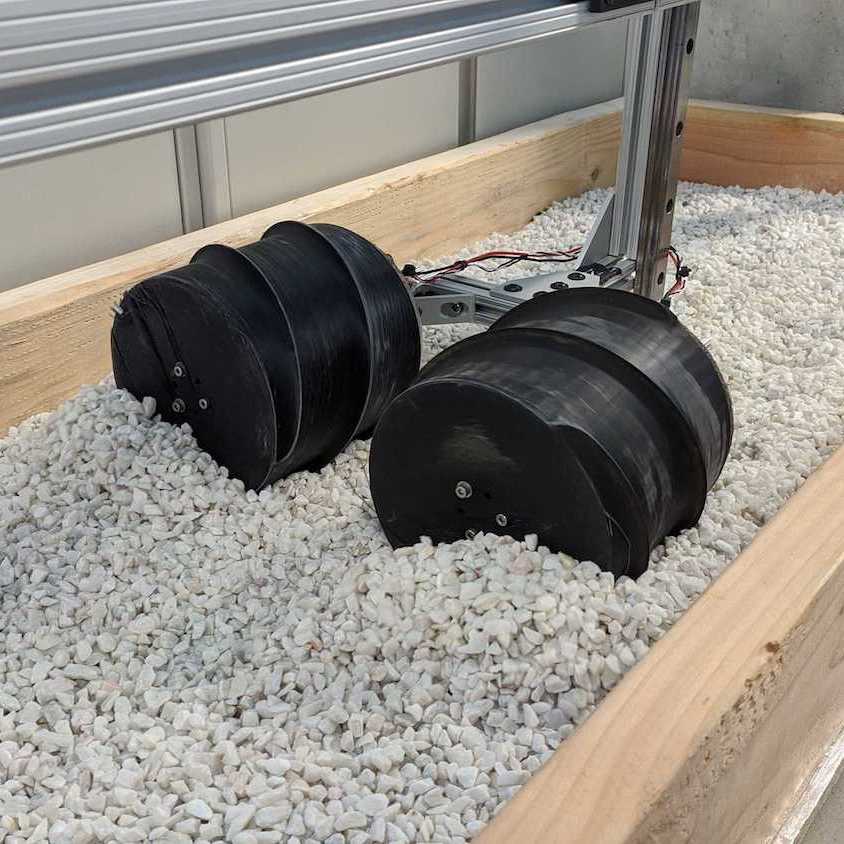}
    \includegraphics[width=0.49\linewidth]{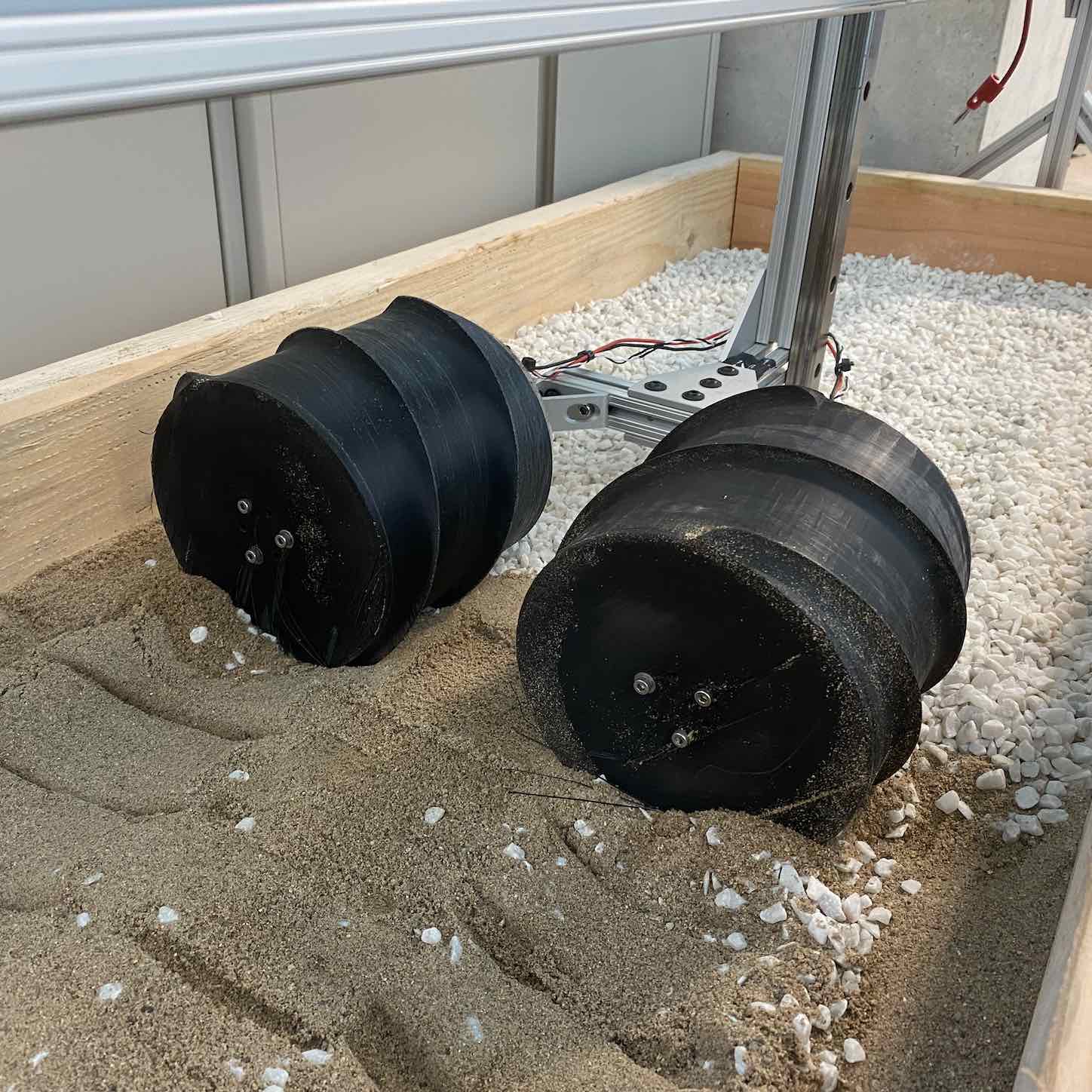}
    \caption{Figures from the parallel configuration experiments where the media is small gravel and small gravel \& sand as shown in the left and right figures, respectively.}
    \label{fig:parallel_experiment_photo}
\end{figure}

\begin{figure}
    \centering
    \includegraphics[trim={1cm 1.5cm 2.5cm 3.5cm},clip, width=\linewidth]{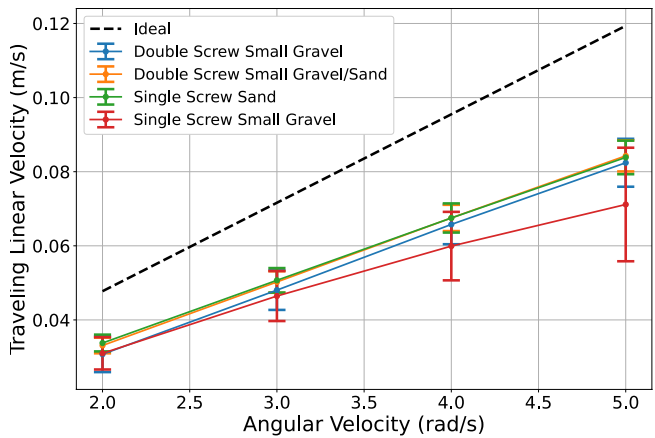}
    \caption{The plots show traveling velocity and longitudinal slip as a function of input angular velocity from our parallel configuration experiment. We also include the single-segment results in the corresponding media to provide perspective about how our experimental results correlate with screw-based locomotion systems.}
    \label{fig:parallel_screw_results}
    \vspace{-5mm} 
\end{figure}

\subsection{Parallel Configuration}

A parallel screw configuration is attached to the test bed as shown in Fig. \ref{fig:parallel_experiment_photo}.
With this modification, the FTS has to be removed as the added mass and torque exceeded the sensor's safe operating range. Therefore we could only collect measurements from the screw and linear axis motors.
The media used for these experiments is small gravel and small gravel \& sand.
The screws are set to regulate a constant velocity of 2, 3, 4, and 5 rad/s.
The measured locomotion velocity and longitudinal slips are shown in Fig. \ref{fig:parallel_screw_results}.
The measured longitudinal slippage and COT are $0.33 \pm 0.07$ and $2.45$, respectively, for the gravel and $0.29 \pm 0.04$ and $2.28$, respectively, for the gravel \& sand.








\section{Discussion}

The static test results highlight two essential properties: a max applied and resultant torque exists when remaining static in different media, and the MA increases with the input torque when remaining static.
This implies that maximal static efficiency is achieved at the transition point directly below the maximum environment-dependent torque.
From the screw velocity plots in Fig. \ref{fig:single_vel}, it is apparent that the input-to-output velocity relationship is linear even with an axial load.
The axial load, however, does increase the longitudinal slippage, thrust in motion, locomotion efficiency, and COT, as seen in Table \ref{tab:media}. The increased thrust in motion is due to the screw working to overcome the axial load. However, producing this extra thrust does not require much additional input torque from the motor; thus, the efficiency also increases. This may be partially explained by the axial load causing the screw to embed better into the media, gaining more traction and generating better propulsion. There is a limit to this effect, however, if the screw digs itself in too much, then overall performance drops significantly, as shown in the sand test. 

As expected, screws cannot provide any locomotion on rigid surfaces such as concrete.
Meanwhile, the more granular media such as gravel and wood chips have the highest locomotion efficiency when under axial load and the lowest COT. We believe this occurs because these media have a higher shearing force and provide better traction.
Sand, on the other hand has a significant drop in locomotive efficiency when an axial load is applied, and we believe this is because the screw digs itself too much into the media, as seen in Fig. \ref{fig:stuck}.
These are crucial phenomena for future screw-based locomotion systems to understand as it implies a limit to the amount of towing ability (axial load) in highly granular environments like sand. Another result of note is not just the average longitudinal slip; the standard deviation indicates how consistently the screw can locomote well. Locomotion on media with a high standard deviation of longitudinal slip, such as gravel, indicates that the screw's velocity is erratic. The gravel's high shearing force and low friction provided a good performance overall, but the erratic motion behavior may be undesirable under specific scenarios. 

The parallel configuration experiments highlight how our results correspond to previously proposed screw-locomotion systems \cite{neumeyer_1965, He_2017}. 
The results line up with the single-segment experiments;
however, a slight performance increase is seen in longitudinal slip and increased COT, likely due to the averaging effect of counter-rotating parallel screws.

The two most influential media properties affecting performance are frictional resistance and shearing strength. High frictional resistance reduces efficiency and mechanical advantage, while a low shearing strength allows the screw to displace media more easily, also leading to reduced efficiency. Another critical factor that is not explored here is the effect of the weight of the screw configuration. Preliminary tests showed that decreasing the vertical normal force significantly reduced the thrust produced and increased the longitudinal slip.



\section{Conclusion}
\begin{figure}
    \centering
    \vspace{2mm}
    \includegraphics[width=0.95\linewidth, trim={0 0 0 3.5cm},clip]{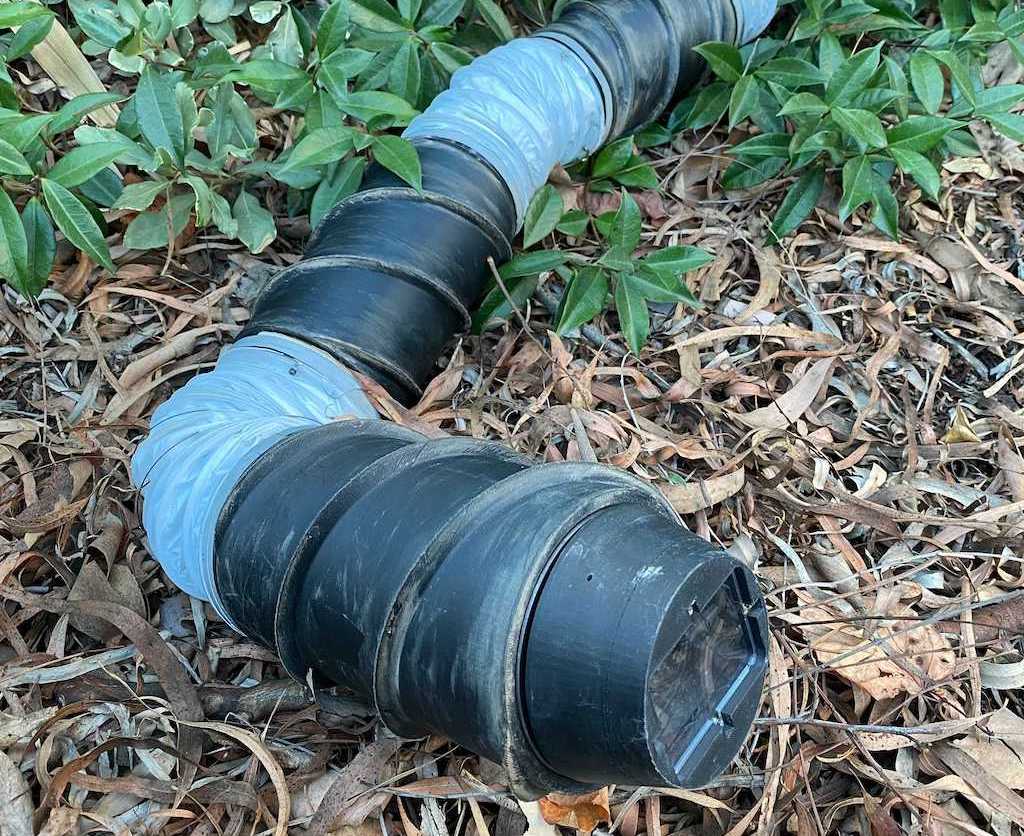}
    \caption{The experimental process and results of screw-locomotion characterization over multiple media is of significance to multi-domain robotic platforms operating autonomously, such as the ARCSnake~\cite{arcsnake_tro} and the EELS successor~\cite{carpenter2021exobiology}, where a wide range of media may be encountered and where communication to a human operator may be all but non-existent.}
    \label{fig:arcsnake}
\end{figure}

This work performed experimental exploration for static and dynamic properties of screw-based locomotion across various test materials, presented novel analysis and metrics for this locomotion mechanism, and designed a test bed providing key functionality required for further characterization.
The experimental results provided several incites key to the future use of screw-based locomotion for multi-terrain mobility.
Within granular material, system power efficiency is maximized by remaining within the boundary between high applied load and slipping.
Intuitively, this may be due to maximizing the usable work used for the task rather than overcoming static systems and environmental losses.
This variable and material-dependent value must be considered in real-world untethered applications. The generated force vectors of a single screw segment are essential for the design and control of robots with variable multi-screw configurations, such as the ARCsnake robot, shown in Fig. \ref{fig:arcsnake}.

Limitations of the current work include the single screw thread pitch, thread size, and screw material.
Additionally, fully fluidic materials and system behavior with the fluidic film caused by the regelation of ice will be explored.

\section{Acknowledgements}

The authors would like to thank Hoi Man (Kevin) Lam and Casey Price for their continued support of the project, and Brett Butler and the Qualcomm Institute Prototyping Lab at UC San Diego for allowing us use of their equipment for 3D printing.
D. Schreiber is supported via the National Science Foundation Graduate Research Fellowship and UC San Diego Accelerating Innovations to Market Grant.
This work was supported by NSF \#1935329.



\clearpage
\balance
\bibliographystyle{ieeetr}
\bibliography{refs}
\balance

\end{document}